\documentclass[sigconf]{acmart}

\AtBeginDocument{%
  }

\setcopyright{acmlicensed}
\copyrightyear{2018}
\acmYear{2018}
\acmDOI{XXXXXXX.XXXXXXX}
\acmConference[Conference acronym 'XX]{Make sure to enter the correct
  conference title from your rights confirmation email}{June 03--05,
  2018}{Woodstock, NY}

\acmISBN{978-1-4503-XXXX-X/2018/06}

\acmSubmissionID{3590}

\usepackage{wrapfig}
\usepackage{multirow}
\usepackage{hyperref}
\usepackage{url}
\usepackage{times}  
\usepackage{helvet} 
\usepackage{courier}  
\usepackage{graphicx}
\usepackage{diagbox}
\usepackage{wrapfig}
\usepackage{subfigure}
\usepackage{graphicx} 
\usepackage{pdflscape} 
\usepackage{booktabs} 
\usepackage{diagbox} 
\usepackage{pifont} 
\begin{document}
\settopmatter{printacmref=false} 
\title{StackCLIP: Clustering-Driven Stacked Prompt in Zero-Shot Industrial Anomaly Detection}

\author{Yanning Hou}
\affiliation{%
  \institution{School of Artificial Intelligence, Anhui University}
  \city{Hefei}
  \country{China}
}
\email{yanning_hou@stu.ahu.edu.cn}
\author{Yanran Ruan}
\affiliation{%
  \institution{School of Artificial Intelligence, Anhui University}
  \city{Hefei}
  \country{China}
}
\email{yanran_ruan@stu.ahu.edu.cn}

\author{Junfa Li}
\affiliation{%
  \institution{School of Artificial Intelligence, Anhui University}
  \city{Hefei}
  \country{China}
}
\email{junfali@stu.ahu.edu.cn}
\author{Shanshan Wang}
\affiliation{%
  \institution{School of Artificial Intelligence, Anhui University}
  \city{Hefei}
  \country{China}
}
\email{wang.shanshan@ahu.edu.cn}

\author{Jianfeng Qiu}
\affiliation{%
  \institution{School of Artificial Intelligence, Anhui University}
  \city{Hefei}
  \country{China}
}
\email{qiujianf@ahu.edu.cn}
\author{Ke Xu}
\authornote{Corresponding author}
\affiliation{%
  \institution{School of Artificial Intelligence, Anhui University}
  \city{Hefei}
  \country{China}
}
\email{xuke@ahu.edu.cn}

\renewcommand{\shortauthors}{Trovato et al.}
\renewcommand\footnotetextcopyrightpermission[1]{}
\begin{abstract}

Enhancing the alignment between text and image features in the CLIP model is a critical challenge in zero-shot industrial anomaly detection tasks. Recent studies predominantly utilize specific category prompts during pretraining, which can cause overfitting to the training categories and limit model generalization. To address this, we propose a method that transforms category names through multicategory name stacking to create stacked prompts, forming the basis of our StackCLIP model. Our approach introduces two key components. The Clustering-Driven Stacked Prompts (CSP) module constructs generic prompts by stacking semantically analogous categories, while utilizing multi-object textual feature fusion to amplify discriminative anomalies among similar objects. The Ensemble Feature Alignment (EFA) module trains knowledge-specific linear layers tailored for each stack cluster and adaptively integrates them based on the attributes of test categories. These modules work together to deliver superior training speed, stability, and convergence, significantly boosting anomaly segmentation performance. Additionally, our stacked prompt framework offers robust generalization across classification tasks. To further improve performance, we introduce the Regulating Prompt Learning (RPL) module, which leverages the generalization power of stacked prompts to refine prompt learning, elevating results in anomaly detection classification tasks. Extensive testing on seven industrial anomaly detection datasets demonstrates that our method achieves state-of-the-art performance in both zero-shot anomaly detection and segmentation tasks.
\end{abstract}

\ccsdesc[500]{Computing methodologies~Visual inspection; Anomaly detection}

\keywords{Anomaly Detection , Zero-shot , CLIP}


\maketitle

\section{Introduction}
Industrial anomaly detection
~\cite{AD1,AD2,AD3,AD4}\begin{figure}[!htb]
    \centering    
 \includegraphics[width=0.48\textwidth]{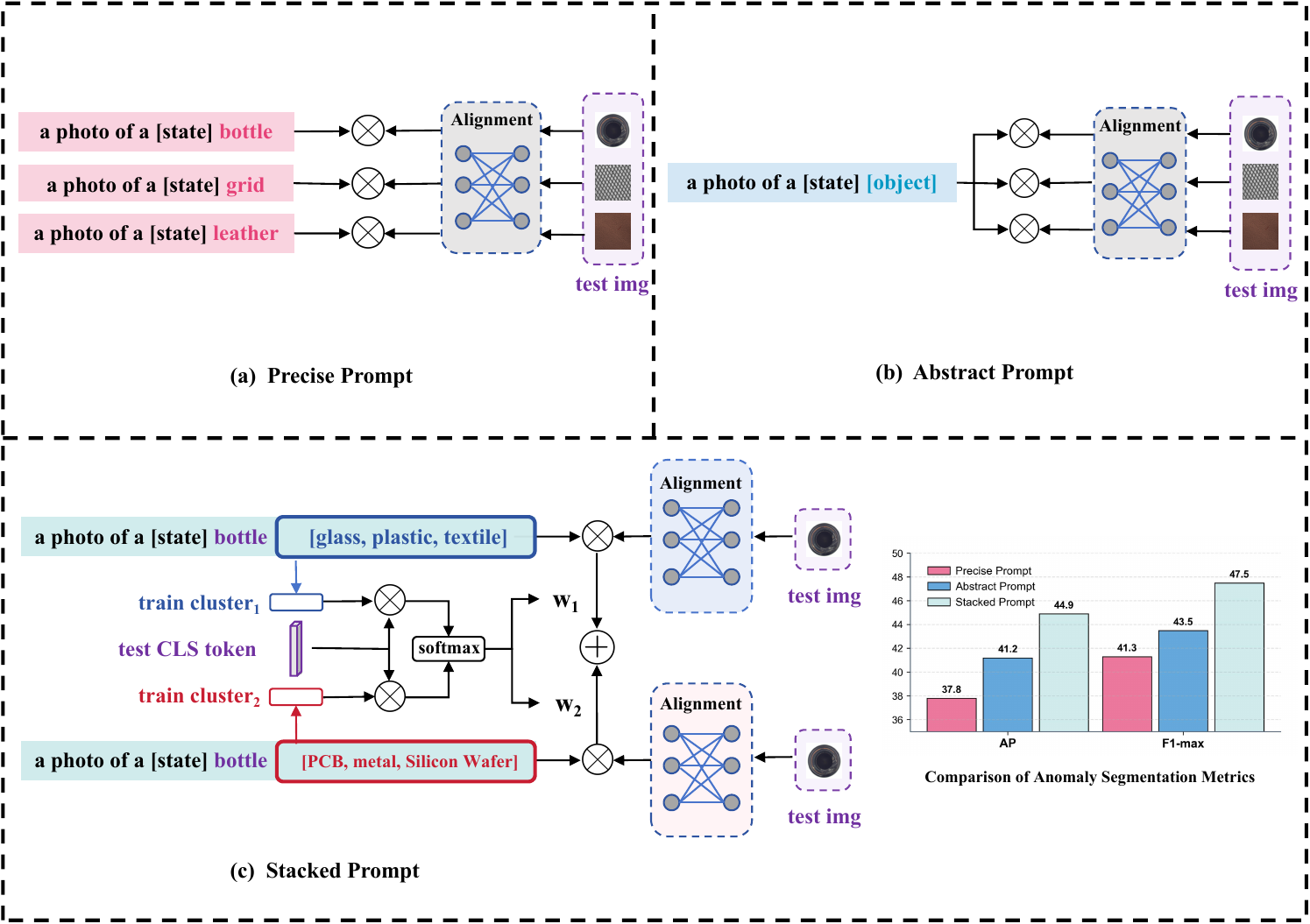} 
    \caption{Comparison of different text prompting methods.  
     (a) Precise prompts  (b) Abstract prompts (c) Stacked prompt}
    \label{fig:fig1}
\end{figure}  involves two main tasks: Anomaly classification (distinguishing normal from anomalous images) and Anomaly segmentation (achieving pixel-level localization). Due to the scarcity of anomalous samples, effective feature extraction is challenging, so traditional approaches typically use unsupervised~\cite{unsupervised1,unsupervised2,unsupervised3,unsupervised4}  or self-supervised~\cite{selfsupervised1,selfsupervised2,selfsupervised3,selfsupervised4} learning to build prototypical features from abundant normal samples and detect anomalies through deviation analysis. However, these methods face two issues: they require processing a large amount of data during training, and detecting multiple categories demands category-specific models, causing the number of models to increase linearly with the number of categories. Consequently, developing a unified zero-shot framework for multi-category adaptation without category-specific retraining is a critical challenge.

Recent advances in zero-shot anomaly detection (ZSAD) using CLIP's vision language pre-training have shown remarkable success~\cite{ZSAD1,Anovl,SAA}. ZSAD methodologies typically quantify the congruence between textual descriptors and visual patterns using cosine similarity analysis; regions that exceed predetermined similarity thresholds are identified as anomalous. With respect to text prompts, previous studies can be broadly categorized into two groups. 

The first group, known as precise prompts, was introduced by WinCLIP~\cite{WinCLIP} through the development of the Compositional Prompt Ensemble (CPE). This approach constructs text prompts using a fixed template that combines a state descriptor and a category name, typically formatted as \texttt{"a photo of a [state][cls]"}. Most subsequent studies have adopted this design—for example, APRIL-GAN~\cite{APRIL-GAN} retains the CPE structure while enhancing the framework with a pre-trained linear projection layer and hierarchical feature aggregation to achieve state-of-the-art performance in anomaly segmentation. Although category-specific prompts are intuitive, they tend to make CLIP models overly concentrate on the target object, which may hinder the detection of subtle anomalous features. Furthermore, as the ZSAD task must generalize across multiple object categories, the use of precise class cues during training increases the risk of overfitting and training instability, ultimately impairing the generalization ability of model.

The second group, known as abstract prompts, was proposed by AnomalyCLIP~\cite{AnomalyCLIP}. This approach introduces an object-agnostic concept by replacing specific category labels with the term “object” and rendering the prompt template learnable. In this formulation, the normal text prompt is represented as \texttt{"[${V}_1$][${V}_2$]...[${V}_E$]
[object]"}, while the anomaly text prompt is expressed as \texttt{"[${W}_1$]
[${W}_2$]...[${W}_E$][damaged][object]"}, where ${V}_i$ and ${W}_i$(i $\in$ 1,...,E) are learnable word embeddings in normality and abnormality text prompt templates, respectively. This approach not only eliminates the need for manually designed templates but also effectively prevents the model from overly focusing on the object itself. However, it has a significant limitation in information capture: replacing specific category names with the generic term “object” results in a complete loss of category-specific information. For instance, objects belonging to the "wood" category may exhibit inherent color inconsistencies. When abstract textual prompts omit the category name, the model is deprived of this critical information, which may lead to the misinterpretation of these intrinsic attributes as anomalies.

To mitigate the limitations inherent in the aforementioned text prompt design, this study proposes a strategy based on stacked prompts. We construct the generic prompt by stacking category names, with the text prompt format defined as \texttt{"a photo of a [state][cls$_1$][cls$_2$]...[cls$_n$]"}. The approach not only reduces the model's excessive focus on the object itself but also effectively preserves the critical information conveyed by the category names. This, in turn, further enhances the model's generalization ability and anomaly detection performance. In Fig.\ref{fig:fig1}, we compared the performance of three textual prompt designs on the more challenging task of anomaly detection segmentation. The results indicate that the stacked prompt design achieved the optimal performance.

To further enhance this paradigm, we propose the Cluster-driven Stacked Prompt (CSP). Specifically, CSP first partitions the categories into several semantically related clusters based on the similarity of their textual features. Then, within each cluster, it constructs a corresponding stacked prompt tailored to the local category group. The Ensemble Feature Alignment (EFA) module subsequently trains cluster-specific, knowledge-aware linear layers for each stack, which are adaptively integrated according to the attributes of the test category. Moreover, we observe that the stacked prompt exhibits robust image-level anomaly detection. Unlike traditional prompt learning methods, we introduce Regulated Prompt Learning (RPL), which employs the stacked prompt as a regularization constraint to improve the model's anomaly classification performance.

Extensive experimentation validates our stack prompt paradigm’s effectiveness in zero-shot anomaly adaptation. Our final framework, \textbf{StackCLIP}, establishes new state-of-the-art performance across diverse zero-shot anomaly benchmarks. Our contributions are summarized as follows:\begin{itemize}
\item  We conducted a comparative analysis of precise and abstract prompts, and introduced a novel stacked prompt paradigm that offers improved adaptability and performance.
\item We proposed Clustering-Driven Stacked Prompt (CSP) and Ensemble Feature Alignment (EFA) modules using stacked prompts for pre-training, enhancing feature alignment, and achieving accurate anomaly segmentation.
\item  We introduced Regulating Prompt Learning (RPL), using stacked prompts to regularize prompt learning and complete anomaly classification.
\item Comprehensive experiments on multiple industrial anomaly detection datasets show that the simple and efficient approach of StackCLIP achieves excellent zero-shot anomaly detection performance on diverse defect detection data.
\end{itemize}
\section{Related work}
\label{sec:formatting}

\subsection{Zero-shot Anomaly Detection.}
In industrial anomaly detection, pre-trained vision models~\cite{VIT,BLIP,CLIP,SAM,DINO} have shown strong performance due to their excellent generalization and feature extraction capabilities. Current anomaly detection methods based on pre-trained large models can be divided into two types:
The first type~\cite{WinCLIP,SAA} requires no additional training, such as WinCLIP~\cite{WinCLIP} and SAA~\cite{SAA}. WinCLIP~\cite{WinCLIP} uses a sliding window method to extract multi-granularity image features for alignment, achieving significant results in classification tasks. However, it requires multiple encodings of the same image to capture anomalous features. SAA~\cite{SAA} combines the capabilities of Grounding DINO~\cite{GroundingDINO} and SAM~\cite{SAM}, using Grounding DINO~\cite{GroundingDINO} for localization through text prompts and SAM~\cite{SAM} for segmentation via box prompts. However, this method has high costs and long inference times.
The second type~\cite{APRIL-GAN,AnomalyCLIP,AdaCLIP,CLIP-SAM,CLIP-SAM2,CLIP-AD} requires additional training on anomaly detection data. APRIL-GAN~\cite{APRIL-GAN} introduced using a linear layer to enhance the alignment of text and image features, succeeding in anomaly segmentation, but it ignored classification tasks. Similarly, SDP~\cite{CLIP-AD} uses a linear layer to strengthen feature alignment and incorporates CLIP Surgery~\cite{CLIP-surgery} with a V-Vattention dual-branch structure. Although this improves anomaly detection, the dual-branch structure increases computational costs.
\subsection{Prompt Learning in Vision-Language Models.}
Prompt Learning~\cite{hou2025soft}, as an efficient alternative to parameter tuning, achieves satisfactory results with fewer tuned parameters compared to traditional full-network fine-tuning. CoOp~\cite{CoOp} introduced learnable text prompts for few-shot classification, and DenseCLIP~\cite{DenseCLIP} expanded this concept to dense prediction tasks by adding an image decoder. PromptSRC~\cite{promptsrc} introduced regularization through raw feature output, while AnomalyCLIP~\cite{AnomalyCLIP} applied prompt learning to industrial anomaly detection, proposing object-agnostic prompts to avoid issues caused by varying object semantics. AnomalyCLIP~\cite{AnomalyCLIP}, with its glocal context optimization, can capture local anomalous semantics, enabling it to perform both classification and segmentation without needing an additional decoder. However, the dual-branch structure in AnomalyCLIP~\cite{AnomalyCLIP} increases model complexity and computational costs. Additionally, the pre-training approach may cause underfitting or overfitting, affecting model stability and generalization.
\section{Approach}
\subsection{Overview}
\begin{figure*}[!htb]
    \centering
    \includegraphics[width=1\textwidth]{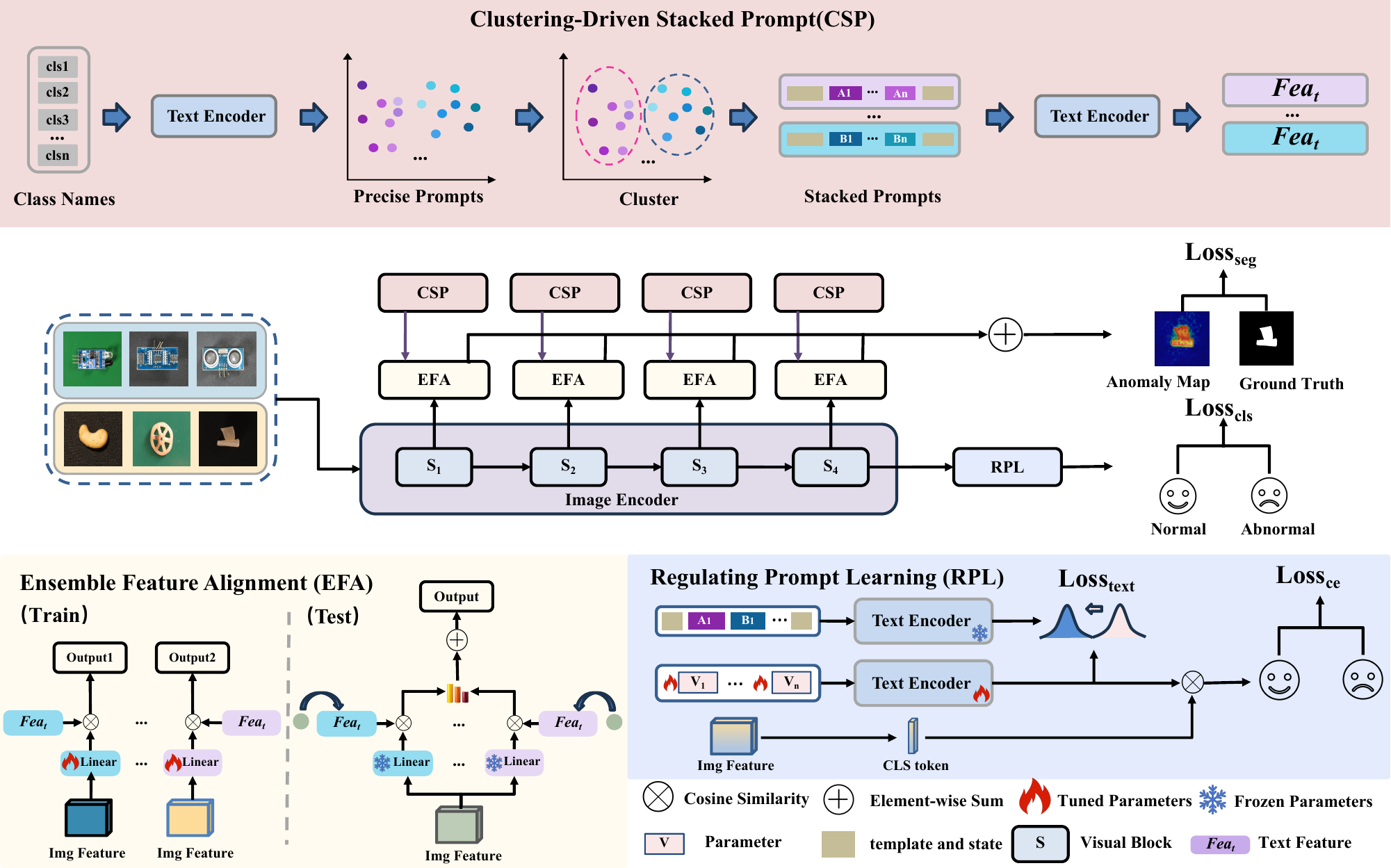} 
    \caption{Overview of StackCLIP To improve image-text feature alignment for anomaly detection, StackCLIP introduces Clustering-Driven Stacked Prompts (CSP) and an Ensemble Feature Alignment (EFA) module to capture diverse anomalous attributes. In classification tasks, it further employs a Regulating Prompt Learning (RPL) module, which regularizes stacked prompts to boost classification accuracy and enhance generalization. }
    \label{fig:fig2}
\end{figure*}

In this paper, we introduce StackCLIP, which improves segmentation and classification performance in industrial anomaly detection through stacked prompts. As shown in Fig.~\ref{fig:fig2}, StackCLIP first uses the Clustering-Driven Stacked Prompt (CSP) module to categorize the training data (see Sec.\ref{sec:CSP}). It then applies the Ensemble Feature Alignment (EFA) module (see Sec.\ref{sec:EFA}) to learn different anomalous features, improving the alignment of image and text features. For anomaly detection classification tasks, we propose the Regulating Prompt Learning (RPL) module, which leverages the broad generalization ability of stacked prompts to regularize prompt learning (see Sec.\ref{sec:RPL}), thus enhancing classification performance.
\subsection{Clustering-Driven Stacked Prompt (CSP)}\label{sec:CSP}

Stacked text prompts form the foundation of our study; their basic form is as follows:
\begin{equation}
\mathcal{S}^p=\texttt{a photo of a [state][cls$_1$][cls$_2$]...[cls$_n$]}
\end{equation}
Here, \texttt{[state]} denotes the text that describes the normal or anomalous condition of object, and \texttt{[cls]} represents the stacked category names.
To determine which category names should be grouped for stacking, we adopt a clustering-based approach. Specifically, all category names \( \mathcal{C} = \{c_1, c_2, \dots, c_n\} \) are embedded into textual feature representations \( T = \{t_1, t_2, \dots, t_n\} \), where each \( t_i  \) is obtained via the CLIP text encoder using precise prompt templates. These features are then clustered using the k-means algorithm.

To select the optimal set of categories for stacking, we introduce a scoring mechanism based on the intra-cluster variance, penalized by the number of clusters:
\begin{equation}
\boldsymbol{n^*}=\arg\min_n\left(\sum_{i=1}^n\sum_{t\in T_i}\frac{1}{|T_i|}\left\|t-t*\right\|^2+\boldsymbol{\lambda(n)} \right)
\end{equation}
In the formula, \(t\) denotes an individual text feature and \(t^*\) represents the centroid of its corresponding cluster, with \(T_i\) being the set of all text features in the \(i\)-th cluster. The term\( \sum_{t \in T_i} \frac{1}{|T_i|} \| t - t^* \|^2 \)computes the average intra-cluster distance, while the penalty term \(\lambda(n) = 0.1 \times e^n\) is introduced to discourage excessive clustering.  The optimal number of clusters \( n^* \) minimizes the overall score, guiding the selection of category sets that both align well semantically and support robust training.
Finally, for each cluster 
\( i \), we define the associated category label for stacking using a prompt that aligns with the stacked prompt structure:
\begin{equation}
\mathcal{S}^p_i=\texttt{a photo of a [state][cluster$_i$]}
\end{equation}
\begin{equation}
\texttt{[cluster$_i$]} =\texttt{[cls$_{i,1}$][cls$_{i,2}$]...[cls$_{i,n}$]}
\end{equation}

This template generates a set of generic text prompts for each cluster, where \texttt{[cluster$_{i}$]} refers to the label of the \( i \)-th cluster.On the other hand,\texttt{[cls$_{i,j}$]} represents the class name of a specific category within the \( i \)-th cluster. Each cluster encompasses categories that carry distinct semantic meanings, allowing the model to acquire corresponding knowledge through these text prompts.

\subsection{Ensemble Feature Alignment (EFA)}\label{sec:EFA}
To better capture the semantic heterogeneity present in industrial visual data, we propose the Ensemble Feature Alignment (EFA) module. This component performs group-specific cross-modal alignment by tightly binding each image cluster to its own stacked prompt representation and projection heads, enabling adaptive multi-level supervision across both spatial and semantic axes.
Based on the clustering results from the CSP module, we partition the complete dataset into \( n^* \) clusters, denoted as
\begin{equation}
\mathcal{I} = \{ \mathcal{I}_1, \mathcal{I}_2, \dots, \mathcal{I}_{n^*} \},
\end{equation}
where \( \mathcal{I}_i \) denotes the \(i\)-th image cluster. Each cluster \( \mathcal{I}_i \) is associated with a unique stacked prompt \( \mathcal{S}_i^p \), forming a strict one-to-one mapping between groups and prompts. This design ensures that each group of images is semantically guided by a tailored prompt, facilitating more precise alignment.

For every image \( I_{i,j} \in \mathcal{I}_i \), we extract multi-level features from the CLIP image encoder at layers \( \ell \in \{6, 12, 18, 24\} \):
\begin{equation}
\mathbf{h}_{i,j}^\ell = \phi^\ell(I_{i,j}),
\end{equation}
where \( \phi^\ell(\cdot) \) denotes the output feature from layer \( \ell \). To align these visual features with the corresponding textual representation, we apply a group-specific linear projection:
\begin{equation}
\tilde{\mathbf{h}}_{i,j}^\ell = \mathbf{W}_{i}^\ell\mathbf{h}_{i,j}^\ell + \mathbf{b}_{i}^\ell,
\end{equation}
where \( \mathbf{W}_{i}^\ell \) and \( \mathbf{b}_{i}^\ell \) are the learnable projection weights and biases specific to group \( i \) and layer \( \ell \). Simultaneously, the stacked prompt \( \mathcal{S}_i^p \) is encoded using the CLIP text encoder:
\begin{equation}
\mathbf{t}_i = \psi(\mathcal{S}_i^p).
\end{equation}

The anomaly evidence is then computed by measuring the cosine similarity between each projected visual feature and the group-specific text feature:
\begin{equation}
M_{i,j}^\ell = \cos(\tilde{\mathbf{h}}_{i,j}^\ell, \mathbf{t}_i) \in \mathbb{R}^{H \times W}.
\end{equation}
Note that each image yields multiple anomaly maps \( \{ M_{i,j}^\ell \} \), one per selected layer. During training, we froze the parameters of CLIP and used focal loss ($\mathcal{L}_{\mathrm{focal}}$) and dice loss ($\mathcal{L}_{\mathrm{dice}}$) functions to optimize the linear layers. This approach aims to improve the alignment between text features and image features, thereby enhancing the performance of anomaly detection segmentation.

\begin{equation}
\begin{aligned}
\mathcal{L}_{\mathrm{focal}} = 
&-\alpha(1-M_{i,j}^\ell)^{\gamma}\log(M_{\mathrm{f}})M_{\mathrm{gt}} \\
&-(1-\alpha)(M_{i,j}^\ell)^{\gamma}\log(1-M_{i,j}^\ell)(1-M_{\mathrm{gt}})
\end{aligned}
\end{equation}

\begin{equation}
\mathcal{L}_{\mathrm{dice}}=1-\frac{2\sum(M_{i,j}^\ell\cdot M_{\mathrm{gt}})+\epsilon}{\sum(M_{i,j}^\ell)+\sum(M_{\mathrm{gt}})+\epsilon}
\end{equation}

where \( M_{\text{gt}} \) is the ground truth anomaly map and the hyperparameters \( \alpha \), \( \gamma \), and \( \epsilon \) are set to 1, 2, and 1, respectively. The final loss function is \( \mathcal{L}_{\text{seg}} = \mathcal{L}_{\text{focal}} + \mathcal{L}_{\text{dice}} \).

During inference, we still employ stacked prompt strategy to enable generalized anomaly detection across both seen and unseen categories. Specifically, we construct multiple sets of textual prompt features by combining the current test category with the training-derived cluster categories. This design allows our model to dynamically adapt to multiple projection layers during testing.
 Given a test image \( I_t \), we first extract the CLS token from the final transformer layer of the CLIP image encoder as a global representation:

\begin{equation}
\mathbf{h}_{\text{cls}} = \phi^{\ell=24}_{\text{cls}}(I_t)
\end{equation}

This CLS token serves as a reference to calculate the alignment strength between the test image and each of the training clusters. For each cluster \(i\) from the training set, we generate a corresponding textual prompt by combining the current test class \(c_t\) with the training cluster label \(i\):
\begin{equation}
\mathcal{S}^p_{t,i}=\texttt{a photo of a [cls$_{test}$][cluster$_i$]}
\end{equation}

Each textual prompt \( \mathcal{S}^p_{t,i} \) is then passed through the CLIP text encoder to generate the corresponding textual embedding \( \mathbf{t}_{t,i} \).

Next, we compute the cosine similarity between the CLS token \( \mathbf{h}_{\text{cls}} \) and the textual embeddings \( \mathbf{t}_{i} \) of each cluster. The cosine similarity is used to obtain attention weights \( \alpha_i \) that reflect the alignment between the test image and each training cluster:

\begin{equation}
\alpha_i = \frac{\exp\left( \cos(\mathbf{h}_{\text{cls}}, \mathbf{t}_{i}) \right)}{\sum_{j=1}^{n^*} \exp\left( \cos(\mathbf{h}_{\text{cls}}, \mathbf{t}_{j}) \right)}
\end{equation}
These weights reflect how well the test image aligns with each learned semantic cluster and are used to modulate the contribution of each layer’s output.

In parallel, we extract multi-level image features from the intermediate layers of the CLIP image encoder, specifically from layers \( \ell \in \{6, 12, 18, 24\} \). These feature maps \( \Phi(I_t) \) contain both spatial and semantic information that is important for anomaly localization. To align the image features with each prompt group, we apply a group-specific linear projection:

\begin{equation}
\tilde{\mathbf{f}}^{\ell}_i = \mathbf{W}^{\ell}_i \cdot \phi^{\ell}(I_t) + \mathbf{b}^{\ell}_i
\end{equation}

For each prompt group \(i\) and layer \( \ell \), we compute the cosine similarity between the projected image features \( \tilde{\mathbf{f}}^{(\ell)}_i \) and the corresponding textual embeddings \( \mathbf{t}_{t,i} \). This results in a raw anomaly map \( M_i^{(\ell)} \) for each combination of image feature layer and prompt group:

\begin{equation}
M_i^{\ell} = \cos\left( \tilde{\mathbf{f}}^{\ell}_i, \mathbf{t}_{t, i} \right)
\end{equation}

Finally, we aggregate the raw anomaly maps \( M_i^{\ell} \) using the attention weights \( \alpha_i \) calculated earlier. The final anomaly detection map \( M_{\text{final}} \) is computed as the weighted sum of all anomaly maps from different prompt groups and image feature layers:
\begin{equation}
M_{\text{final}} = \sum_{\ell \in \{6,12,18,24\}} \sum_{i=1}^{n^*} \alpha_i \cdot M_i^{\ell}
\end{equation}

\subsection{Regulating Prompt Learning (RPL)}\label{sec:RPL}

To enhance the generalization capability of the classification branch in our anomaly detection framework, we introduce a prompt regularization strategy based on stacked prompts. This design leverages the inherent generalization power of prompt-based vision-language models and ensures that the learned prompts remain semantically meaningful and robust to category shifts.

In this module, we adopt a dual-objective training approach that combines supervised classification loss with a regularization loss. Specifically:
\begin{table*}[htbp]
  \centering
  \caption{Performance comparison of SOTA approaches on the MVTec-AD~\cite{MVTec} and VisA~\cite{visa} datasets. Evaluation metrics include AUROC, {$F_1$-max}, AUPRO, and AP. Bold indicates the best performance and underline indicates the runner-up.}
    \resizebox{0.99\textwidth}{!}{
    \begin{tabular}{ccccccccccc}
    \toprule
    \multirow{2}[4]{*}{Dataset} & \multirow{2}[4]{*}{Method} & Supervised  & \multicolumn{3}{c}{Pixel-level} &       & \multicolumn{3}{c}{Image-level} & \multirow{2}[4]{*}{Rank} \\
\cmidrule{4-6}\cmidrule{8-10}          &       & Traning & AUPRO & AP    & $F_1$-max &       & AUROC & AP    & $F_1$-max &  \\
    \midrule
    \multirow{7}[2]{*}{\textbf{MVTec-AD}~\cite{MVTec}} & WinCLIP~\cite{WinCLIP} & \ding{55} & 64.6  & 18.2  & 31.7  &       & \underline{91.8}  & \underline{96.5}  & \underline{92.9}  & 4.2 \\
          & APRIL-GAN~\cite{APRIL-GAN} & \ding{51}     & 44.0  & \underline{40.8}  & \underline{43.3}  &       & 86.1  & 93.5  & 90.4  & 4.7 \\
          & CLIP Surgery~\cite{CLIP-surgery} &  \ding{55} & 69.9  & 23.2  & 29.8  &       & 90.2  & 95.5  & 91.3  & 5.2 \\
          & SAA+~\cite{SAA}  & \ding{55} & 42.8  & 37.8  & 28.8  &       & 63.1  & 81.4  & 87.0  & 6.3 \\
          & SDP+~\cite{CLIP-AD}  & \ding{51}     & \underline{85.1}  & 36.3  & 40.0  &       & \textbf{92.2} & \textbf{96.6} & \textbf{93.4} & \underline{2.0} \\
          & AnomalyCLIP~\cite{AnomalyCLIP} & \ding{51}     & 81.4  & 34.5  & 39.1  &       & 91.5  & \textbf{96.6} & 92.7  & 3.3 \\
          & \textbf{StackCLIP(ours)} & \ding{51}     & \textbf{86.4} & \textbf{46.0} & \textbf{47.6} &       & 91.7  & \textbf{96.6} & 92.7  & \textbf{1.7} \\
    \midrule
    \multirow{7}[2]{*}{\textbf{VisA}~\cite{visa}} & WinCLIP~\cite{WinCLIP} & \ding{55} & 56.8  & 5.4   & 14.8  &       & 78.1  & 81.2  & 79.0  & 5.3 \\
          & APRIL-GAN~\cite{APRIL-GAN} & \ding{51}     & 86.8  & \underline{25.7}  & \underline{32.3}  &       & 78.0  & 81.4  & 78.7  & 3.5 \\
          & CLIP Surgery~\cite{CLIP-surgery} & \ding{55} & 64.7  & 10.3  & 15.2  &       & 76.8  & 80.2  & 78.5  & 5.8 \\
          & SAA+~\cite{SAA}  & \ding{55} & 36.8  & 22.4  & 27.1  &       & 71.1  & 77.3  & 76.2  & 5.8 \\
          & SDP+~\cite{CLIP-AD}  & \ding{51}     & 83.0  & 18.1  & 24.6  &       & 78.3  & 82.0  & 79.0  & 3.8 \\
          & AnomalyCLIP~\cite{AnomalyCLIP} & \ding{51}     & \underline{87.0}  & 21.3  & 28.3  &       & \underline{82.1}  & \underline{85.4}  & \underline{80.4}  & \underline{2.5} \\
          & \textbf{StackCLIP(ours)} & \ding{51}     & \textbf{89.8} & \textbf{28.0} & \textbf{34.2} &       & \textbf{84.7} & \textbf{86.9} & \textbf{82.7} & \textbf{1.0} \\
    \bottomrule
    \end{tabular}%
    }
  \label{tab:2}%
\end{table*}%
\begin{itemize}
    \item \textbf{Classification Loss} ($\mathcal{L}_{\mathrm{ce}}$): We use the standard cross-entropy loss to supervise the classification of image features aligned with the prompt representations. This encourages the model to correctly identify normal versus anomalous categories.
    \item \textbf{Prompt Regularization Loss} ($\mathcal{L}_{\mathrm{text}}$): To constrain the learned prompts and prevent overfitting, we introduce a mean squared error (MSE) loss that aligns the prompt representation with a fixed text embedding derived from a stacked prompt constructed using training cluster information. This alignment helps preserve the generalizable semantics across seen and unseen categories.
\end{itemize}

Formally, the two loss functions are defined as:
\begin{equation}
\mathcal{L}_{\mathrm{ce}} = - y\log(p_y), \quad
\mathcal{L}_{\mathrm{text}} = \frac{1}{d} \sum_{m=1}^{d} (t{'}_m - t^{s}_m)^2
\end{equation}

where $p_y$ is the predicted probability for the true class label $y$, computed based on the cosine similarity between the test image’s CLS token ${h}_{\text{cls}}$ and the learned prompt embedding $t'$. $d$ denotes the dimensionality of the textual feature vector, $t'$ is the learnable prompt embedding, and $t^s$ is the reference embedding obtained from a pre-defined stacked prompt during training.

The final training objective combines both components: $\mathcal{L}_{\text{cls}} = \mathcal{L}_{\mathrm{ce}} + \mathcal{L}_{\mathrm{text}}$To simplify the training procedure, we use a stacked prompt with a single cluster (i.e., cluster number = 1). This shared prompt is sufficient to guide the classification process. During inference, this fixed set of learned prompts is reused for all categories, enabling efficient and scalable anomaly classification.

\section{Experiments}

\subsection{Experimental Setup}
We conducted a series of experiments to assess the anomaly detection performance of our method in a zero-shot setting, focusing on the latest and most challenging industrial anomaly detection benchmarks. We also performed extensive ablation studies to validate the effectiveness of each component we proposed.
\paragraph{Datasets and Metrics.}

We conduct experiments on seven real industrial datasets, including MVTec-AD~\cite{MVTec}, VisA~\cite{visa}, DTD-Synthetic~\cite{DTD-Synthetic},
, MPDD~\cite{MPDD}, DAGM~\cite{DAGM}, SDD~\cite{KSDD} and BTAD~\cite{BTAD}.
We conducted a fair and comprehensive comparison with existing zero-shot anomaly detection and segmentation (ZSAS) methods using widely adopted metrics, namely AUROC, AP, AUPRO, and {$F_1$-max}. The anomaly detection performance is evaluated using the Area Under the Receiver Operating Characteristic Curve (AUROC). AP quantifies the accuracy of the model at different recall levels. The PRO metric represents the coverage of the segmented region over the anomalous region. {$F_1$-max} represents the harmonic mean of precision and recall at the optimal threshold, implying the model's accuracy and coverage. 

\paragraph{Implementation Details.}
\begin{figure*}[t] 
    \centering
    \begin{minipage}[b]{0.695\textwidth} 
        \includegraphics[width=\textwidth]{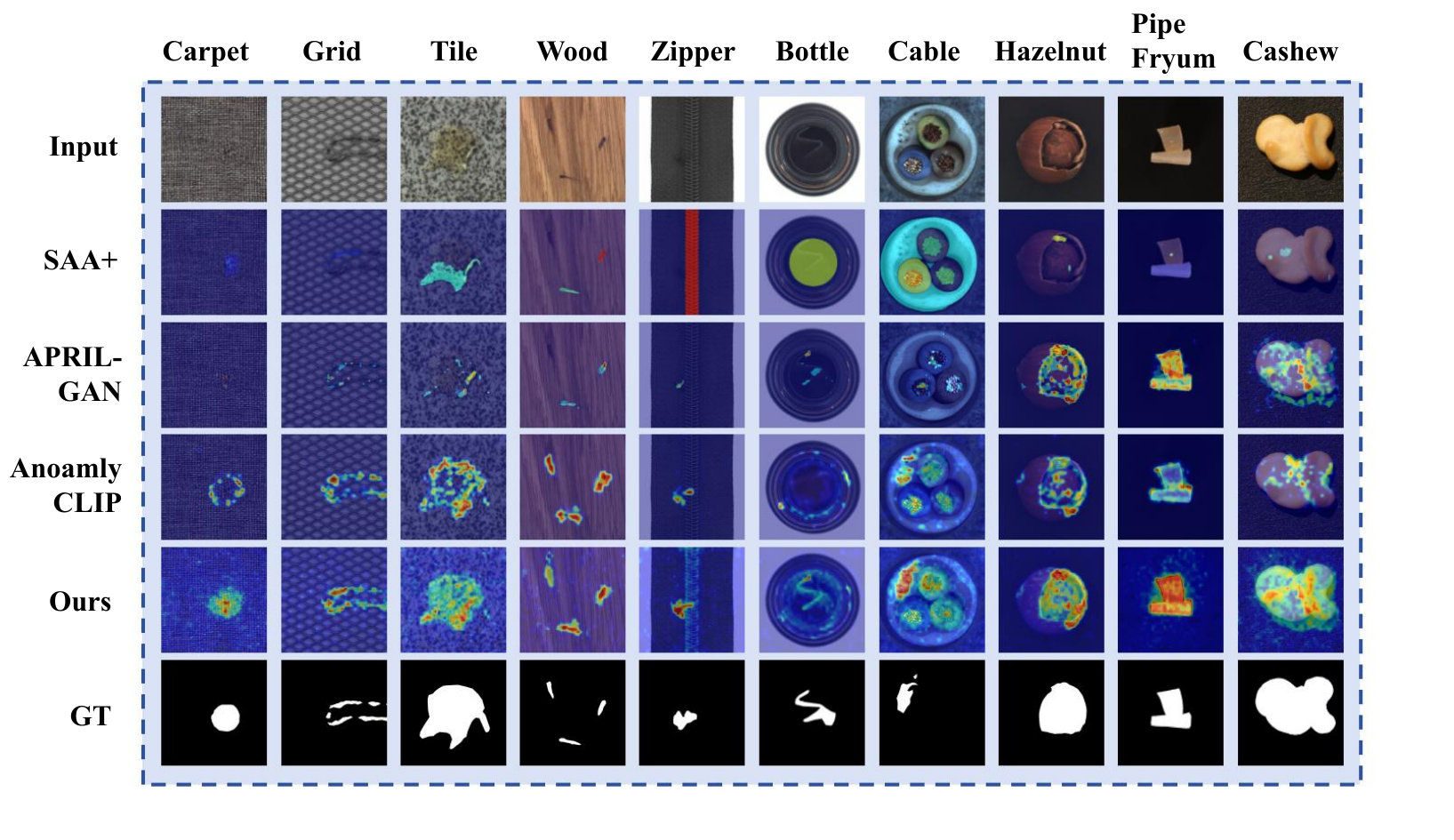}
        \captionof{subfigure}{Comparison of zero-shot anomaly segmentation.} 
        \label{fig:vis1}
    \end{minipage}%
    \begin{minipage}[b]{0.305\textwidth} 
        \includegraphics[width=\textwidth]{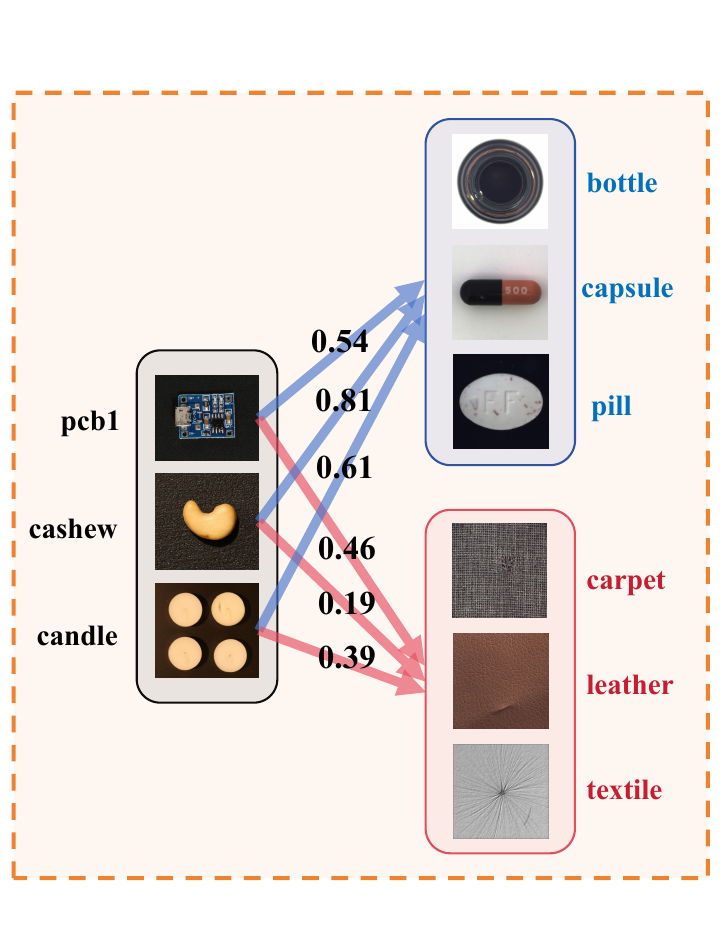}
        \captionof{subfigure}{Test data selection across linear layers for anomaly alignment}
        \label{fig:vis2}
    \end{minipage}
    \caption{Visual Comparison of Zero-Shot Anomaly Segmentation and Test Data Selection Patterns}
    \label{fig:fig3}
\end{figure*}
Due to the ZSAD task, it is necessary to ensure that the auxiliary data does not contain the content of the test dataset. Specifically, when testing on MVTec-AD~\cite{MVTec}, we use the training set of VisA~\cite{visa}, and conversely, when testing on VisA, we use the training set of MVTec-AD~\cite{MVTec}. We use the publicly available CLIP model (ViT-L/14@336px) as our backbone and extract patch embeddings from the 6-th, 12-th, 18-th, and 24-th layers. The images used for training and testing are scaled to a resolution of 518 × 518. All parameters of the CLIP model are frozen. The framework training employ the Adam optimizer. For the linear training sets MVTec-AD~\cite{MVTec} and VisA~\cite{visa}, the learning rates are set at 1e-4 and 1e-3 for this stage, respectively, while in the prompt learning stage, both are set at 1e-4. Training proceeds for 2 epochs with a batch size of 16. We set the length of the learnable word embeddings to 12. These learnable token embeddings are appended to the first 9 layers of the text encoder to refine the text space, with each layer having a length of 20. All experiments were conducted using PyTorch 2.4.0 and run on a single NVIDIA RTX 3090 24GB GPU.

\begin{table*}[htbp]
  \centering
  \caption{Pixel-level performance (AUPRO, AP, $F_1$-max) across merged datasets with varying cluster number. }
  \renewcommand{\arraystretch}{1.5}
  \resizebox{\textwidth}{!}{%
    \begin{tabular}{>{\centering\arraybackslash}p{10em}>{\centering\arraybackslash}p{6.135em}ccccccccc}
    \toprule[\heavyrulewidth] 
    \multicolumn{2}{c}{\multirow{2}[2]{*}{\diagbox[width=13em,height=2.0em]{\textbf{\large Train}}{\textbf{\large Test}}}} & \multirow{3}[4]{*}{\textbf{\Large MVTec}~\cite{MVTec}} & \multirow{3}[4]{*}{\textbf{\Large VisA}~\cite{visa}} & \multirow{3}[4]{*}{\textbf{\Large DTD}~\cite{DTD-Synthetic}} & \multirow{3}[4]{*}{\textbf{\Large MPDD}~\cite{MPDD}} & \multirow{3}[4]{*}{\textbf{\Large DAGM}~\cite{DAGM}} & \multirow{3}[4]{*}{\textbf{\Large BTAD}~\cite{BTAD}} & \multirow{3}[4]{*}{\textbf{\Large SDD}~\cite{KSDD}} & \multirow{3}[4]{*}{\textbf{\Large Average}} \\
    \multicolumn{2}{c}{} &       &       &       &       &       &       &       &  \\
    \cmidrule{1-2}    
    \multicolumn{1}{>{\centering\arraybackslash}p{10em}}{\textbf{\normalsize Train Data}} & \multicolumn{1}{>{\centering\arraybackslash}p{6.135em}}{\textbf{\normalsize Cluster Num}} &       &       &       &       &       &       &       &  \\
    \hline
   \multirow{2}[1]{*}{\textbf{\large VisA}} & \textbf{1} & (\textbf{86.6},\underline{44.2},\underline{46.6}) & ( —   ,   —   ,   — ) & (\underline{90.7},\underline{66.0},\textbf{65.2}) & (\underline{88.5},\underline{28.6},\underline{33.5}) & (\textbf{80.2},\underline{44.3},\underline{48.6}) & (\textbf{79.3},\underline{47.7},\underline{51.8}) & (\textbf{88.1},\underline{14.4},\underline{27.6}) & (\textbf{85.6},\underline{40.9},\underline{45.6}) \\
          & \textbf{2} & (\underline{86.4},\textbf{46.0},\textbf{47.6}) & ( —  ,  —  ,  — ) & (\textbf{91.3},\textbf{66.4},\underline{65.1}) & (\textbf{89.1},\textbf{31.2},\textbf{35.1}) & (\underline{79.3},\textbf{45.9},\textbf{50.0}) & (\underline{76.0},\textbf{48.6},\textbf{53.5}) & (\underline{88.0},\textbf{22.1},\textbf{35.4}) & (\underline{85.0},\textbf{43.4},\textbf{47.8}) \\
    \hline
    \multirow{2}[0]{*}{\textbf{\large MVTec+MPDD}} & \textbf{1} & ( —  ,  —  ,  — ) & (\textbf{89.4},\textbf{25.7},\textbf{32.0}) & (\textbf{92.5},\underline{68.4},\underline{66.9}) & ( —  ,  —  ,  — ) & (\textbf{81.1},\underline{48.3},\underline{52.1}) & (\underline{73.5},\underline{37.1},\underline{42.4}) & (\underline{93.7},\underline{41.9},\underline{47.3}) & (\underline{86.0},\underline{44.3},\underline{48.1}) \\
    & \textbf{2} & ( —  ,  —  ,  — ) & (\underline{88.9},\underline{22.8},\underline{30.0}) & (\underline{91.9},\textbf{70.3},\textbf{68.9}) & ( —  ,  —  ,  — ) & (\underline{79.4},\textbf{49.6},\textbf{53.4}) & (\textbf{76.1},\textbf{40.8},\textbf{44.4}) & (\textbf{94.1},\textbf{44.2},\textbf{50.1}) & (\textbf{86.1},\textbf{45.5},\textbf{49.4}) \\
    \hline
    \multirow{3}[0]{*}{\textbf{\large MVTec+DTD}} & \textbf{1} & ( —  ,  —  ,  — ) & (87.8,\underline{23.6},\underline{30.4}) & ( —  ,  —  ,  — ) & (85.9,\textbf{27.4},\textbf{33.4}) & (77.8,49.3,54.6) & (68.7,\underline{32.0},\underline{37.2}) & (\underline{89.6},\textbf{44.1},\textbf{51.0}) & (82.0,\underline{35.3},\underline{41.3}) \\
          & \textbf{2} & ( —  ,  —  ,  — ) & (\underline{88.4},20.5,27.5) & ( —  ,  —  ,  — ) & (\textbf{88.3},\underline{26.6},\underline{33.3}) & (\underline{79.7},\textbf{54.0},\textbf{57.0}) & (\underline{72.0},28.0,33.9) & (\textbf{92.9},39.3,48.4) & (\underline{84.3},33.7,40.0) \\
          & \textbf{3} & ( —  ,  —  ,  — ) & (\textbf{89.1},\textbf{23.9},\textbf{30.8}) & ( —  ,  —  ,  — ) & (\underline{87.6},26.5,33.0) & (\textbf{80.6},\underline{53.3},\underline{56.1}) & (\textbf{73.4},\textbf{32.7},\textbf{37.5}) & (\textbf{92.9},\underline{41.5},\underline{50.3}) & (\textbf{84.7},\textbf{35.6},\textbf{41.5}) \\
    \hline
     \multirow{3}[1]{*}{\textbf{\large MVTec+DAGM}} & \textbf{1} & ( —  ,  —  ,  — ) & (\underline{90.0},\underline{22.7},\underline{30.1}) & (\underline{94.2},\underline{75.6},\underline{73.6}) & (89.5,\underline{27.2},\underline{34.1}) & ( —  ,  —  ,  — ) & (78.9,42.4,47.7) & (93.2,41.0,\underline{51.5}) & (89.2,\underline{41.8},47.4) \\
       & \textbf{2} & ( —  ,  —  ,  — ) & (\textbf{90.1},21.5,\underline{30.1}) & (\underline{94.2},71.9,72.9) & (\underline{89.7},26.2,32.4) & ( —  ,  —  ,  — ) & (\underline{81.2},\underline{45.7},\underline{50.3}) & (\textbf{95.4},\underline{42.4},\textbf{51.6}) & (\underline{90.1},41.5,\underline{47.5}) \\
     & \textbf{3} & ( —  ,  —  ,  — ) & (\textbf{90.1},\textbf{23.1},\textbf{30.4}) & (\textbf{94.9},\textbf{76.6},\textbf{74.0}) & (\textbf{90.4},\textbf{28.1},\textbf{34.8}) & ( —  ,  —  ,  — ) & (\textbf{81.7},\textbf{}46.1,\textbf{}51.6) & (\underline{95.3},\textbf{42.8},50.1) & (\textbf{90.5},\textbf{43.3},\textbf{48.2}) \\
    \bottomrule[\heavyrulewidth] 
    \end{tabular}%
  }
  \label{tab:6}%
\end{table*}
\subsection{Performance Comparison with SOTA Method}
\begin{figure*}[!htb]
    \centering
    \includegraphics[width=1\textwidth]{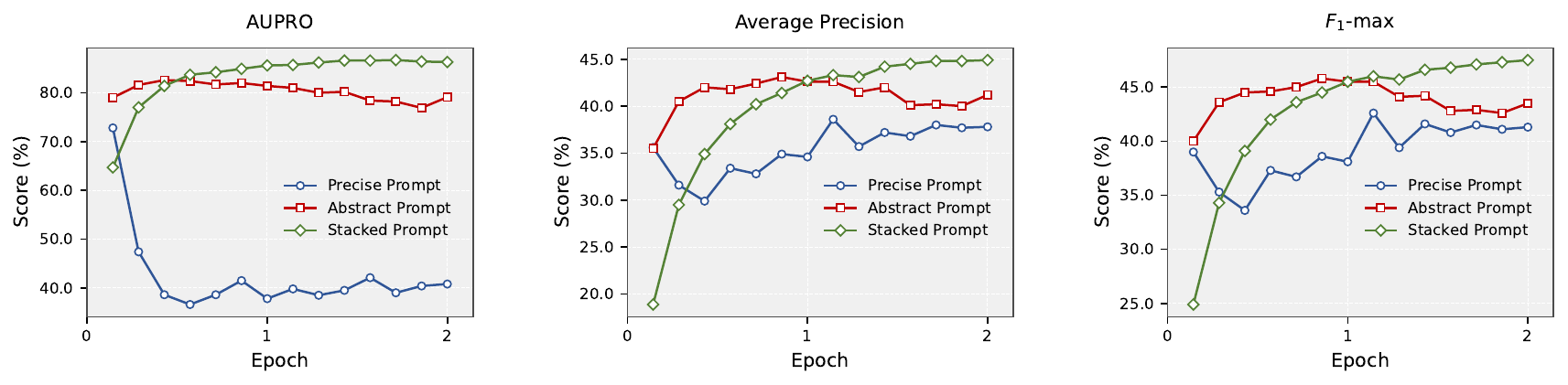} 
    \caption{Performance comparison of different prompt strategies across evaluation metrics}
    \label{fig:compare}
\end{figure*}
We compared methods without the need for training, such as WinCLIP~\cite{WinCLIP}, SAA~\cite{SAA}, and CLIP Surgery~\cite{CLIP-surgery}, to those that necessitate training, including APIRL-GAN~\cite{APRIL-GAN}, CLIP-AD~\cite{CLIP-AD}, AnomalyCLIP~\cite{AnomalyCLIP}. We use the experimental results from the original paper, and since the CLIP Surgery~\cite{CLIPSurgery} and AnomalyCLIP~\cite{AnomalyCLIP} methods only have some metrics, we reproduce the results using the original code and the weight files provided in the code.

As shown in Table~\ref{tab:2}, StackCLIP achieves state-of-the-art performance across both segmentation and classification tasks under the zero-shot setting. On MVTec-AD~\cite{MVTec}, StackCLIP outperforms prior methods in segmentation, achieving an AUPRO of 86.4, 1.3 higher than SDP+~\cite{CLIP-AD}, and improving AP and $F_1$-max by 9.7 and 7.6, respectively. For classification, it reaches 91.7 AUROC and 92.7 $F_1$-max, only 0.5 and 0.7 lower than SDP+'s 92.2 and 93.4, respectively. On VisA~\cite{visa}, StackCLIP surpasses all baselines across all metrics. Compared to SDP+~\cite{CLIP-AD}, it improves AUROC by 6.4 and $F_1$-max by 3.7, and exceeds AnomalyCLIP~\cite{AnomalyCLIP} by 2.6 in AUROC. The performance gap between datasets can be attributed to their characteristics: MVTec-AD~\cite{MVTec} features simpler, localized anomalies where dual-branch designs like SDP+~\cite{CLIP-AD} are effective; VisA presents diverse and cluttered anomalies where StackCLIP’s stacked prompt design and ensemble feature alignment offer superior generalization and robustness.

In Fig.~\ref{fig:fig3} (a), we present visual results of zero-shot anomaly segmentation (ZSAS) to further validate the effectiveness of our proposed method. We also compare our approach with other methods such as SAA+~\cite{SAA}, APRIL-GAN~\cite{APRIL-GAN}, SDP+~\cite{CLIP-AD}, and AnomalyCLIP~\cite{AnomalyCLIP}. As WinCLIP~\cite{WinCLIP} is not open-source, it was not included in our comparison
In comparison to these methods, our approach demonstrates stronger performance in both localization and segmentation of anomaly regions, providing more accurate identification of anomalous areas and yielding superior segmentation results. We further present in Fig.~\ref{fig:fig3} (b) the selection patterns of test data across different linear layers, which highlights the superior capacity of the EFA module to capture diverse anomaly-related features and to adaptively select the most informative representations. This demonstrates its effectiveness in enhancing feature alignment and model adaptability in the presence of varying anomaly types.

\subsection{Ablation Studies}
To validate the effectiveness of our method, we conducted a component-wise analysis of the module in our framework.

\paragraph{The Effectiveness of Stacked Prompt}
To align text and image features effectively, we evaluate the impact of different prompt types under the setting where only a set of linear projection heads~\cite{APRIL-GAN} is trained. We evaluate the differences among various prompt types in terms of training stability, convergence speed, and final performance after convergence. All models are trained on the VisA~\cite{visa} dataset and evaluated on the MVTec-AD~\cite{MVTec} dataset. All plotted curves are based on the testing results from MVTec-AD~\cite{MVTec}.

The performance of the three prompt types precise prompts, abstract prompts, and stacked prompts is presented in Fig.~\ref{fig:fig1}, with their training dynamics illustrated in Figure~\ref{fig:compare}. Compared to the precise prompt, the stacked prompt achieves improvements of 42.6, 3.4, and 3.3 in AUPRO, AP, and $\mathrm{F_1}$-max, respectively. Relative to the abstract prompt, the corresponding gains are 3.4, 2.2, and 2.4.In terms of training behavior, the precise prompt demonstrates substantial oscillations and severe overfitting. For instance, its AUPRO score drops sharply from 72 to 38 within a single epoch. While the abstract prompt converges rapidly, it also exhibits overfitting in the later stages of training. A comparative analysis of the training progression metrics reveals that the stacked prompt consistently outperforms both the precise and abstract prompts across all epochs. It demonstrates superior stability, higher $\mathrm{F_1}$-max scores, enhanced average precision, and significantly more robust AUPRO. 
\begin{figure*}[!htb] 
    \centering
    \includegraphics[width=1\linewidth]{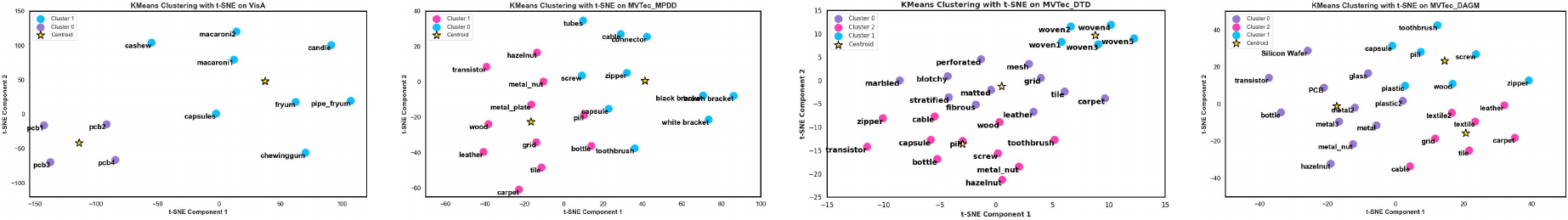}
    \caption{Clustering results of the VisA and three mixed datasets after the Clustering-Driven Stacked Prompt (CSP) module.}
    \label{fig:fi5}
\end{figure*}

\paragraph{The Effectiveness of CSP and EFA }
In our framework, the Clustering-Driven Stacked Prompt (CSP) and Ensemble Feature Alignment (EFA) modules work closely together to accomplish anomaly segmentation tasks. Specifically, in the CSP module, we use the K-means clustering method to select and determine the number of categories, with this process facilitated by our designed scoring mechanism.

\begin{table}[htbp]
    \centering
    \setlength{\tabcolsep}{2.0pt}  
    \caption{The impact of different prompts and data settings on pixel-level results.}
    \begin{tabular}{cccccc}  
    \hline
    \multirow{2}[1]{*}{\textbf{Prompt}} & {\textbf{Data}} & \multicolumn{3}{c}{\textbf{Pixel-level}} & \textbf{\#Imgs} \\
          &    \textbf{Settings}   & \textbf{AUPRO} & \textbf{AP} & \textbf{\boldmath$F_1$\unboldmath-max} & \textbf{Num} \\
    \hline
    Precise & All Data & 44.0  & 40.8  & 43.3  & 2162 \\
    \hline
    \multirow{4}[2]{*}{Stacked} & All Data & \textbf{ 86.6 } & 44.2  & \underline{46.6}  & 2162 \\
          & Cluster$_1$ & 86.1  & \underline{44.3}  & \underline{46.6}  & 1360 \\
          & Cluster$_2$ & 84.1  & 39.9  & 43.4  & 802 \\
          & EFA & \underline{86.4}  & \textbf{ 46.0 } & \textbf{ 47.6 } & 2162 \\
    \hline
    \end{tabular}
    \label{tab:5}
\end{table}

When training on the VisA dataset, the results were surprising: the CSP module successfully divided the dataset into two clusters. Cluster$_2$ includes PCB1, PCB2, PCB3, and PCB4, while Cluster$_1$ contains the remaining categories. We trained and tested these clusters separately using the corresponding linear layers. As shown in Table~\ref{tab:5}, despite significantly reduced training data, the test results for both clusters were similar to those from precise prompt training on the full dataset. Notably, under the Cluster$_1$ setting, the AP metric even slightly exceeded our baseline results.

This suggests that, even with a reduced training set, the CSP module's clustering strategy can effectively improve model performance, especially when the data distribution is clear. Clustering simplifies and aggregates the data, helping the model focus on more representative features, which improves anomaly segmentation.

In addition, the EFA module further enhances performance. It dynamically assigns weights to the linear layers trained through cluster$_1$ and cluster$_2$ based on the distance between the test samples and the cluster centers. This adaptive weighting mechanism ensures that each linear layer exerts a different influence according to its performance in specific categories, effectively integrating the knowledge learned from multiple linear layers to achieve optimal performance.
We believe that the limited sample size and number of categories in the VisA~\cite{visa} dataset impose certain restrictions on the performance of our method. To further validate the effectiveness of our method, we have decided to merge multiple datasets~\cite{DAGM,DTD-Synthetic,MPDD} compare them based on the framework we proposed. As shown in Table~\ref{tab:6}, we also provide the corresponding clustered results in Fig.~\ref{fig:fi5}. With an increase in the number of clusters, after passing through the EFA module, integrating knowledge from different groups with multiple linear layers, almost all evaluation metrics have shown improvement under the same amount of data. However, we observe that in some cases, the performance when using two clusters is slightly lower than that with a single cluster. Our analysis suggests that this counterintuitive drop is mainly caused by significant data shifts introduced by the K-means clustering process, which subsequently affects the training and fitting of the linear layers.

\paragraph{The Effectiveness of RPL}
In our framework, the Regulated Prompt Learning (RPL) module enhances anomaly classification performance by regularizing with stacked prompts. As a first step, we evaluate the classification performance of precise, abstract, and stacked prompts without performing prompt learning.
\begin{table}[htbp]
  \centering
  \caption{Classification Performance Metrics}
  \resizebox{0.45\textwidth}{!}{
  \begin{tabular}{ccccc}
    \hline
    \multirow{2}[2]{*}{\textbf{Method}} & \multicolumn{3}{c}{\textbf{Image-Level}} & \multirow{2}[2]{*}{\textbf{Feature Dim}} \\
          & \textbf{AUROC} & \textbf{AP} & \textbf{\boldmath$F_1$\unboldmath-max} & \\
    \hline
    Precise Prompt  & 86.1  & 93.5  & 90.4  & \phantom{0}$\mathbb{R}^{n \times 2 \times C}$ \\
    Abstract Prompt & 86.4  & 94.3  & 90.8  & \phantom{0}$\mathbb{R}^{1 \times 2 \times C}$ \\
    Stacked Prompt  & \textbf{87.7} & \textbf{94.6} & \textbf{90.9} & \phantom{0}$\mathbb{R}^{1 \times 2 \times C}$ \\
    \hline
  \end{tabular}%
  }
  \label{tab:cls}%
\end{table}%
As shown in Table~\ref{tab:cls}, even without prompt learning, stacked prompts already exhibit superior performance compared to precise and abstract prompts across all three classification metrics on the MVTec-AD dataset. Specifically, the stacked prompt achieves an AUROC of 87.7, AP of 94.6, and an F1-max of 90.9, outperforming both precise and abstract prompt variants. Compared to the precise prompt ($\mathbb{R}^{n \times 2 \times C}$), which assigns individualized prompts to each sample, and the abstract prompt ($\mathbb{R}^{1 \times 2 \times C}$), which uses a single general prompt for all inputs, the stacked prompt maintains the generality of the abstract prompt while introducing greater semantic diversity. 

\begin{table}[htbp]
  \centering
  \caption{RPL module ablation study}
    \begin{tabular}{cccc}
    \toprule
    \multirow{2}[4]{*}{\textbf{Setting}} & \multicolumn{3}{c}{\textbf{Image-Level}} \\
\cmidrule{2-4}          & \textbf{AUROC} & \textbf{AP} & \textbf{$F_1$-max} \\
    \midrule
     $\mathcal{L}_{\mathrm{ce}}$& 83.4  & 89.5  & 90.5 \\
    $\mathcal{L}_{\mathrm{ce}}+\mathcal{L}_{\mathrm{text}}$& \textbf{91.7} & \textbf{96.6} & \textbf{92.7} \\
    \bottomrule
    \end{tabular}%
  \label{tab:7}%
\end{table}%
Table~\ref{tab:7} presents the ablation results of our RPL module. When using only the classification loss $\mathcal{L}{\mathrm{ce}}$, the model achieved an AUROC of 83.4, an AP of 89.5, and an $F_1$-max of 90.5. It is worth noting that this result is even lower than the one obtained by directly using precise prompts, indicating that, without additional constraints, prompt learning is highly prone to overfitting. After incorporating the regularization loss $\mathcal{L}{\mathrm{text}}$ with stacked prompts, the AUROC increased by 8.3 points to 91.7, the AP improved by 7.1 points to 96.6, and the $F_1$-max rose by 2.2 points to 92.7. These results demonstrate that the introduction of the stacked prompt regularization loss substantially improves the model's image-level anomaly classification performance. We attribute this improvement to the enhanced generalization ability provided by the regularization mechanism. Additionally, the effectiveness of prompt learning is influenced by several key factors, including the depth (M) and length (L) of the learnable token embeddings, the number of prompts (E), and the initialization strategy. A detailed analysis of these factors is provided in the Appendix.
\section{Conclusion}
\label{conclusion}
In this paper, we introduce the concept of stacked prompts and apply it to industrial anomaly detection. This pre-training method is not only simple and efficient but also significantly enhances the classification and segmentation capabilities in anomaly detection. Furthermore, as the amount of data and the number of categories increase, our method can continuously improve anomaly detection performance in real industrial environments by introducing additional linear layers, allowing for anomaly detection from different perspectives. However, our method still has some limitations. Specifically, its performance may not meet expectations for categories that are difficult to accurately describe with text. Although the EFA module can integrate multiple forms of knowledge, we perform well when the clustering boundaries are clear but encounter difficulties in situations where the boundaries are ambiguous. In future work, we will further explore methods to enhance the alignment capabilities of CLIP features to address more complex scenarios in industrial anomaly detection.


\bibliographystyle{ACM-Reference-Format}
\bibliography{sample-base}

\appendix

\end{document}